%% file: main_acl.tex
\documentclass[11pt]{article}

\usepackage[preprint]{acl}

\usepackage{times}
\usepackage{latexsym}

\usepackage[T1]{fontenc}
\usepackage[utf8]{inputenc}

\usepackage{microtype}

\usepackage{inconsolata}

\usepackage{graphicx}

\usepackage{booktabs}
\usepackage{multirow}
\usepackage{multicol}
\usepackage{makecell}
\usepackage{svg}
\usepackage{tcolorbox}
\usepackage{longtable}
\tcbuselibrary{breakable}
\usepackage{hyperref}
\usepackage{amsmath} 
\title{ARC: Active and Reflection-driven Context Management for Long-Horizon Information Seeking Agents}

\author{
 \textbf{Yilun Yao\textsuperscript{1}},
 \textbf{Shan Huang\textsuperscript{2}},
 \textbf{Elsie Dai\textsuperscript{1}},
 \textbf{Zhewen Tan\textsuperscript{1}},
\\
 \textbf{Zhenyu Duan\textsuperscript{3}},
 \textbf{Shousheng Jia\textsuperscript{3}},
 \textbf{Yanbing Jiang\textsuperscript{1}},
 \textbf{Tong Yang\textsuperscript{1}},
\\
\\
 \textsuperscript{1}Peking University,
 \textsuperscript{2}Beihang University,
 \textsuperscript{3}Qiyuan Tech,
\\
}
\newcommand{\ARC}{ARC}

\begin{document}
\maketitle

\input{Body/0.abstract}
\input{Body/1.intro}

\input{Body/2.related}
\input{Body/3.methodology}
\input{Body/4.experiments}

\input{Body/5.conclusion}
\input{Body/limitations}

\bibliography{Body/reference}
\newpage
\input{Body/appendix}

\end{document}

%% file: Body/0.abstract.tex
\begin{abstract}
    Large language models are increasingly deployed as research agents for deep search and long-horizon information seeking, yet their performance often degrades as interaction histories grow.
    This degradation, known as context rot, reflects a failure to maintain coherent and task-relevant internal states over extended reasoning horizons.
    Existing approaches primarily manage context through raw accumulation or passive summarization, treating it as a static artifact and allowing early errors or misplaced emphasis to persist.
    Motivated by this perspective, we propose \textbf{ARC}, which is the first framework to systematically formulate context management as an active, reflection-driven process that treats context as a dynamic internal reasoning state during execution.
    ARC operationalizes this view through reflection-driven monitoring and revision, allowing agents to actively reorganize their working context when misalignment or degradation is detected.
    Experiments on challenging long-horizon information-seeking benchmarks show that ARC consistently outperforms passive context compression methods, achieving up to an \textbf{11\%} absolute improvement in accuracy on BrowseComp-ZH with Qwen2.5-32B-Instruct.
\end{abstract}

%% file: Body/1.intro.tex
\section{Introduction}

Deep search\footnote{Deep search is a class of long-horizon information-seeking tasks that require iterative exploration, multi-step reasoning, and evidence integration over extended interaction trajectories.} and long-horizon information seeking are increasingly important as large language models (LLMs) are applied to real-world research, analysis, and decision-making tasks, motivating the development of deep research systems~\citep{openai2025deepresearch}.
Unlike single-turn question answering, these tasks require agents to explore unfamiliar information spaces over many steps, repeatedly issue search queries, integrate heterogeneous evidence, and sustain coherent reasoning across extended interaction horizons~\citep{chen-etal-2017-reading,yang2018hotpotqa,das2018multistep,wei2025browsecompsimplechallengingbenchmark,mialon2023gaiabenchmarkgeneralai}.

Despite strong reasoning and information-seeking capabilities, LLM performance often degrades in long-horizon settings.
As interaction histories grow, models struggle to maintain coherent and task-relevant internal states, a phenomenon known as \emph{context rot}~\citep{hong2025context}.
Prior work attributes this degradation to challenges in sequential modeling, including long-term credit assignment, representational bottlenecks, and attention dilution in long contexts~\citep{rot-credit-assignment,rot-gradient-zhou2025lengthinducedembeddingcollapseplmbased,rot-bottleneck-capacity}.
These issues are especially pronounced in information-seeking tasks, where agents must integrate heterogeneous evidence across many steps while preserving critical intermediate decisions.
Although modern LLMs support increasingly large context windows~\citep{openai_gpt4.1_2025, meta_llama4_2025}, effectively leveraging such capacity for stable multi-step reasoning remains difficult, underscoring the need for principled \emph{context engineering}.

Most existing approaches to long-horizon information seeking adopt one of two context management strategies.
The first directly appends raw interaction histories—including reasoning, actions, and observations—into the working context~\citep{yao2023react}, leading to rapid context growth and attention dilution.
The second relies on passive compression, periodically summarizing past interactions to control context length~\citep{ReSum}.
While effective at managing budget constraints, this approach treats context as a static storage artifact rather than an actively maintained reasoning state.
Once compressed, early errors, outdated assumptions, or misaligned emphasis are difficult to correct, as prior summaries are rarely revisited.
Together, both strategies share a common limitation: context is primarily managed to satisfy length constraints.
Information, once written—either verbatim or summarized—is seldom reassessed in light of downstream reasoning outcomes.
This suggests that the challenge of long-horizon information seeking lies not only in deciding what to retain, but in enabling context to evolve as the agent’s understanding changes.

\begin{figure}[h]
    \centering
    \hspace*{-0.07\columnwidth}
    \includegraphics[width=1.12\columnwidth]{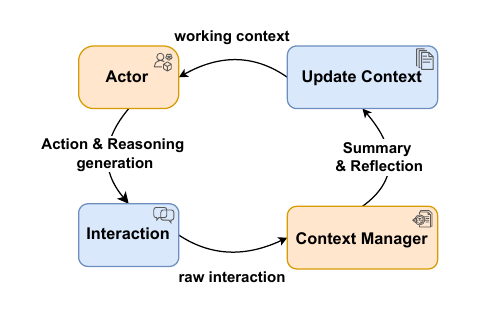}
    \caption{High-level illustration of active and reflection-driven context management in \ARC{}. At each turn, the Context Manager actively updates the working context, enabling the actor to adapt its reasoning as understanding evolves across turns.}
    \label{fig:recon_intro}
\end{figure}

Rather than viewing context as an append-only record or a passively compressed summary, we argue that it should be treated as a dynamic internal state that can be continuously monitored and actively managed.
Motivated by this perspective, we introduce \textbf{\ARC{}}, an active and reflection-driven context management framework that explicitly separates action execution from context management, enabling agents to incrementally construct, revise, and realign their internal context over long interaction horizons (Figure~\ref{fig:recon_intro}).

\ARC{} introduces a new perspective on context management for long-horizon information seeking.
By modeling context as a dynamically evolving internal state that is continuously summarized, monitored, and revised through reflection, \ARC{} enables agents to actively repair degraded context and realign their internal state as new evidence is acquired.

Building on this perspective, we make the following contributions:
\begin{itemize}
    \item We identify a fundamental gap between \emph{passive context compression} and \emph{active context management} in long-horizon information-seeking agents, showing that context management is not merely about fitting history into a limited window, but about continuously maintaining a task-aligned internal reasoning state.
    \item We propose \textbf{\ARC{}}, an active and reflection-driven framework that treats context as a dynamically managed internal state, enabling continuous revision and realignment during reasoning.
    \item We introduce a dual-component agent architecture with a dedicated Context Manager, decoupled from action execution and responsible for online context construction and reflection-driven revision.
\end{itemize}

%% file: Body/2.related.tex
\section{Related Work}

\subsection{Context Management via Accumulation and Summarization}

To handle long interaction histories within limited context windows, researchers have developed strategies to organize and compress information effectively.

Foundational systems draw inspiration from operating systems; for instance, MemGPT~\citep{memGPT} implements an OS-like paging mechanism to manage hierarchical storage based on relevance.
MemoryBank~\citep{MemorBankenhancinglargelanguage} models memory decay via the Ebbinghaus curve, while Mem0~\citep{Mem0} and A-MEM~\citep{Amem} utilize dynamic graphs or triplets. These structures enable querying complex entity relationships beyond simple semantic similarity, improving retrieval coherence.
Parallel research focuses on optimizing context density.
Memory-R1~\citep{MemoryR1} employs reinforcement learning for CRUD-style memory operations to explicitly update information.
Similarly, ReSum~\citep{ReSum} and AgentFold~\citep{AgentFold} treat context compression as an optimization problem, utilizing gradient-based or folding techniques to distill extensive histories into compact representations within bounded budgets.
Recently, MemEvolve~\citep{MemEvolve} moved beyond fixed designs by introducing a meta-evolutionary framework. Using a dual-loop mechanism, it dynamically optimizes the memory architecture itself, automatically restructuring encoding, storage, and retrieval modules.

These works collectively establish a robust foundation for storage efficiency and long-term architectural adaptation.
In this work, we explore a complementary direction: focusing on the \emph{real-time revision} of the working context during the reasoning process itself.

\subsection{Reflection and Self-Correction in Agent Systems}

Reflection mechanisms have become a standard paradigm for enhancing agent reliability.
A prominent line of work focuses on short-term, action-level correction.
Systems like Reflexion~\citep{Reflexionlanguageagentsverbal} and Self-Refine~\citep{Selfrefineiterativerefinementselffeedback} use iterative verbal critiques to guide subsequent attempts in a ``trial-and-error'' loop.
To ground these critiques, frameworks like CRITIC~\citep{criticlargelanguagemodels} augment the process with external verification tools, ensuring self-correction is based on executable evidence.

Another stream applies reflection to accumulate long-term wisdom.
ReasoningBank~\citep{ReasoningBankscalingagentselfevolving} synthesizes successful trajectories into generalized reasoning patterns for future retrieval.
ACE~\citep{agenticcontextengineeringevolving} employs multi-agent debate to evolve persistent ``playbooks,'' while Evo-Memory~\citep{evomemorybenchmarkingllmagent} updates structured memory stores based on task outcomes to refine high-level policies.

Existing literature has thus extensively covered immediate action refinement and cross-task policy evolution.
Our work aims to bridge these levels by investigating how reflection can be applied to maintain and reorganize the \emph{execution-time context} to support ongoing reasoning.

%% file: Body/3.methodology.tex
\section{Methodology}
\label{sec:method}

We propose \textbf{Active and Reflection-Driven Context Management (\ARC{})}, a framework for improving long-horizon agent performance through continuous interaction-memory construction and reflection-driven internal state management.
\ARC{} adopts a \emph{dual-component architecture} that explicitly separates action execution from context management.
It operates at the level of execution-time internal state rather than action generation.
We refer to the learnable model responsible for context management as the \textbf{Context Manager (CM)}.

\ARC{} is motivated by the observation that failures in long-horizon tasks often arise from how interaction history is summarized, organized, and maintained over time, rather than from insufficient reasoning capability.
Instead of treating context as an append-only record or a fixed summary, \ARC{} models it as a \emph{dynamic internal memory state} that is continuously constructed, monitored, and reorganized to support decision making throughout execution.

\subsection{Overview}

\ARC{} decomposes agent behavior into explicit internal representations and cleanly separates action execution from context management.
At each turn, the agent maintains three core internal components:
\begin{itemize}
    \item \textbf{Checklist}: an inspectable and mutable control structure that tracks the agent’s current objectives and execution progress.
    \item \textbf{Interaction Memory (IM)}: a dynamically managed internal memory that compactly summarizes task-relevant information from past interactions.
    \item \textbf{Previous Interaction}: the complete interaction from the preceding turn, including the agent’s internal reasoning, executed action, and resulting observation.
\end{itemize}

Action execution and context management are decoupled.
During execution, the actor generates an internal reasoning state and an action conditioned on the current interaction memory, checklist, and the most recent raw interaction.
In parallel, a dedicated Context Manager incrementally updates the interaction memory through online summarization.
When execution degrades or inconsistencies are detected, the CM enters a reflection phase that jointly reorganizes the interaction memory and revises the checklist.
Crucially, reflection operates solely on internal representations and does not directly generate actions.

\subsection{Active and Reflection-driven Context Management}
\label{sec:context_management}

\begin{figure}[t]
    \centering
    \includegraphics[width=1.05\columnwidth]{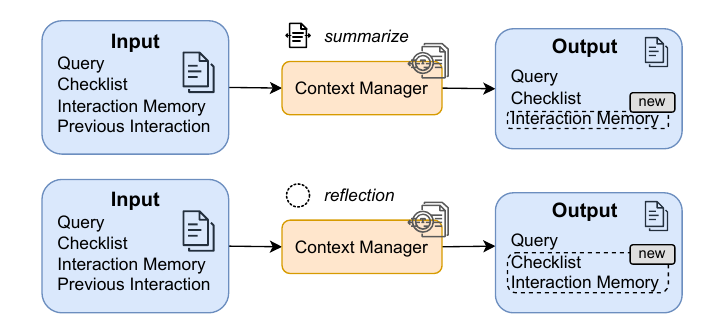}
    \caption{Overview of the Context Manager in \ARC{}.
The upper panel illustrates standard summarization, which compresses accumulated interactions into a compact interaction memory.
The lower panel shows reflection-driven context revision, where the Context Manager actively reorganizes the checklist and interaction memory to correct errors or misalignment, producing an updated execution-time context.}
    \label{fig:recon-cm}
\end{figure}

\ARC{} treats interaction memory as a dynamically managed execution-time internal state.
Rather than serving as a static summary or passive record of past interactions, the interaction memory is continuously updated and, when necessary, actively reorganized to remain aligned with the agent’s evolving understanding of the task.

As illustrated in Figure~\ref{fig:recon-cm}, context management in \ARC{} proceeds at every turn through two tightly coupled processes: incremental summarization and reflection-driven reorganization.

\subsubsection{Active Interaction Memory Construction}

Long-horizon tasks require agents to integrate information across many interaction steps while avoiding unbounded growth of raw interaction history.
To this end, \ARC{} employs an \emph{active, always-on, incremental interaction memory construction mechanism} based on evidence-preserving summarization.

At each turn $t$, after the interaction from the previous turn becomes available, the Context Manager updates the interaction memory according to
\begin{equation}
M_t = \mathrm{Summarize}(Q, L_{t-1}, M_{t-1}, I_{t-1}),
\end{equation}
where $Q$ denotes the task query, $L_{t-1}$ is the checklist state from the previous turn, $M_{t-1}$ represents the interaction memory summarizing all interactions prior to turn $t-1$, and $I_{t-1} = (r_{t-1}, a_{t-1}, o_{t-1})$ is the complete interaction from turn $t-1$, consisting of the agent’s internal reasoning state, executed action, and resulting observation.

This update is strictly incremental.
During normal execution, the Context Manager incorporates information only from the most recent interaction and does not revise, reorder, or remove previously stored memory content.
The summarization operator removes redundant or verbose information while preserving task-relevant evidence and factual fidelity, without introducing new assumptions or speculative inferences.
As a result, the interaction memory maintains a compact yet semantically faithful representation of the agent’s past experience.

Importantly, the interaction from the most recent turn is not immediately compressed at the time of action selection.
Instead, it is retained in raw form and provided directly to the actor at the subsequent turn, ensuring access to fine-grained details before being absorbed into the interaction memory through summarization.

\subsubsection{Reflection-Based Reorganization}

While incremental summarization controls memory growth and preserves semantic coverage, it does not guarantee alignment with the agent’s evolving beliefs, goals, or evidence.
Early incorrect assumptions, outdated information, or misplaced emphasis may therefore persist even under faithful summarization.

To address this limitation, \ARC{} incorporates an active and reflection-driven reorganization mechanism that can be invoked at any turn.
Reflection triggering is integrated into the summarization process: during each incremental update, the Context Manager performs a lightweight self-assessment of the current interaction memory to detect signals of degradation, such as stalled progress, unresolved inconsistencies, or ineffective repetition.
Based on this assessment, the Context Manager decides whether reflection is required.

When triggered, the Context Manager performs a joint internal state update:
\begin{equation}
(M_t, L_t) \leftarrow \mathcal{R}(Q, L_{t-1}, M_t, I_{t-1}),
\end{equation}
where $\mathcal{R}(\cdot)$ denotes a reflection operator that revises both the interaction memory and the checklist conditioned on the task query, prior control state, current memory, and the most recent interaction.

In contrast to incremental summarization, reflection enables non-local modifications of the interaction memory.
The Context Manager may selectively rewrite or merge memory elements to correct errors, remove outdated assumptions, or re-emphasize task-relevant information.
The checklist is revised as part of the same process, updating control-level information such as completed objectives, priorities, or refined subgoals.

Crucially, reflection functions purely as an internal state transformation.
It does not prescribe actions, generate external outputs, or modify the actor’s policy.
By coupling always-on incremental summarization with selectively triggered reflection, \ARC{} actively maintains a coherent and task-aligned interaction memory throughout long-horizon execution.

\subsection{Agent Execution Loop}
\label{sec:loop}

\begin{figure*}[t]
    \centering
    \includegraphics[width=1.02\linewidth]{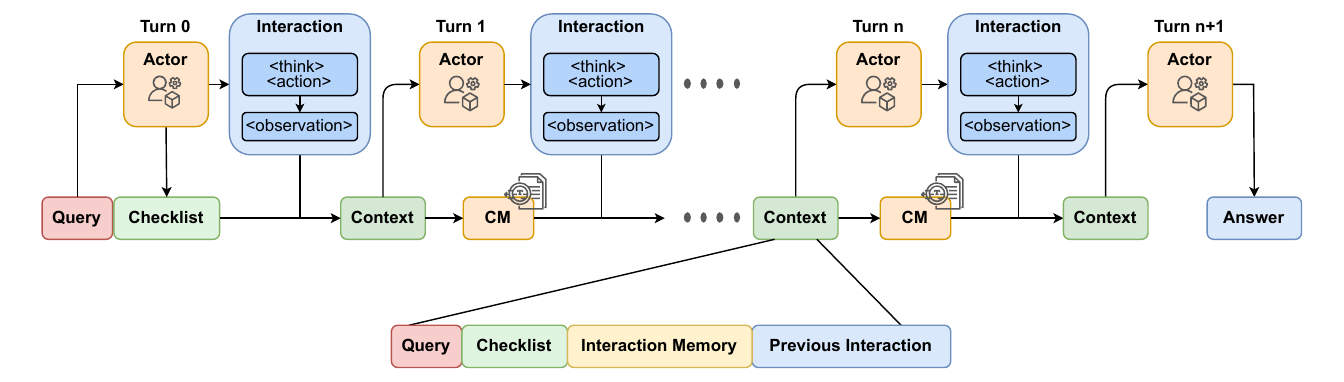}
    \caption{The execution loop of \ARC{}. At each turn, the actor generates reasoning and actions conditioned on the current context, producing an interaction consisting of thoughts, actions, and observations.
    The Context Manager (CM) then integrates the query, checklist, interaction memory, and the most recent interaction to update the context via active context management, which is passed to the next turn until a final answer is produced.}
    \label{fig:agent-loop}
\end{figure*}

As shown in Figure~\ref{fig:agent-loop}, \ARC{} operates in an iterative loop with decoupled action execution and context management.
At each turn $t$, the actor generates an internal reasoning state and an action according to
\begin{equation}
(r_t, a_t) \sim \pi_{\text{actor}}\big(Q,\; L_{t-1},\; M_{t-1},\; I_{t-1}\big),
\end{equation}
where $Q$ denotes the task query, $L_{t-1}$ is the current checklist state, $M_{t-1}$ is the interaction memory summarizing all interactions prior to turn $t$, and $I_{t-1}$ is the complete interaction from the previous turn retained in raw form.

The environment executes $a_t$ and returns a new observation $o_t$.
In parallel, the Context Manager incrementally summarizes $I_{t-1}$ to update the interaction memory.
If reflection is triggered, the Context Manager further reorganizes the interaction memory and revises the checklist, without generating actions.
The loop terminates once a confident answer is reached.

\subsection{Training the Context Manager}
\label{sec:training}

\ARC{} treats context management as a \emph{learnable component} within a dual-component agent architecture.
Accordingly, the Context Manager (CM) is trained to actively maintain and revise interaction memory through incremental summarization and reflection-driven reorganization, independently of action generation.

We train the CM using supervised fine-tuning (SFT) on trajectories collected from information-seeking and search-based tasks.
Each training example consists of an interaction $I_t$ paired with target interaction memory and checklist states $(M_t^*, L_t^*)$ that reflect effective context management behavior.
Because such trajectories require coordinated execution and reflection, we construct a data collection pipeline that leverages strong language models to generate candidate trajectories, followed by filtering to retain only those exhibiting coherent summarization and valid reflection behavior.

The resulting dataset enables the CM to internalize active context management strategies in a single forward pass, replacing fragile prompt-based procedures with robust learned behavior.
Additional details on data collection, and training settings are provided in Appendix~\ref{app:training}.

%% file: Body/4.experiments.tex
\section{Experiments}

\subsection{Experiment Setup}

\paragraph{Datasets}

We evaluate our approach on five benchmarks that require iterative information-seeking with multi-step reasoning: HotpotQA~\citep{yang2018hotpotqa}, GAIA~\citep{mialon2023gaiabenchmarkgeneralai}, xBench-DeepSearch~\citep{chen2025xbenchtrackingagentsproductivity}, BrowseComp~\citep{wei2025browsecompsimplechallengingbenchmark}, and BrowseComp-ZH~\citep{zhou2025browsecompzhbenchmarkingwebbrowsing}. 
For HotpotQA, we randomly sample 512 questions and evaluate agents under an active search-and-browse setting rather than oracle retrieval. 
For GAIA, we use the text-only subset. 
For BrowseComp, we adopt subsets from BrowseComp-LongContext, while executing all tasks under an information-seeking agent setting with iterative search and page inspection. 
All other benchmarks are evaluated on their full test sets.

\paragraph{Baselines}

We compare our method against two representative baselines for context management in information-seeking agents:
\begin{itemize}
    \item \textbf{ReAct}~\citep{yao2023react}directly concatenates the full raw interaction history at each step, including internal reasoning, actions, and observations, without any explicit context compression or revision.
    \item \textbf{ReSum}~\citep{ReSum} performs passive, static summarization to control context length. When the interaction history approaches the maximum context budget, past interactions are compressed into a single summary.
\end{itemize}

All baselines share the same actor model, tool interface and search mechanism.
The only difference lies in how past interactions are represented and maintained in the agent’s internal state.
Additional baseline variants and implementation details are deferred to the Appendix~\ref{app:eval}.

\paragraph{Models}

We evaluate our approach using multiple actor models, including Qwen2.5-7B-Instruct, Qwen2.5-32B-Instruct~\cite{qwen2.5}, Qwen3-14B, Qwen3-32B~\citep{qwen3technicalreport}, and DeepSeek-v3.2~\citep{deepseekai2025deepseekv32}.
The selected actor models span different model families and capacity scales, providing a diverse testbed for evaluating the robustness of \ARC{}.
Unless otherwise specified, all ablation studies and analysis experiments use Qwen2.5-32B-Instruct as the actor model.

Across all settings, the same Context Manager is instantiated using GPT-OSS-120B~\citep{openai2025gptoss120bgptoss20bmodel}.
This design follows the dual-component architecture of \ARC{}, decoupling action generation from context management and ensuring that performance differences primarily arise from the proposed context management strategy rather than disparities in underlying model capacity.

\subsection{Main Results}

Table~\ref{tab:main_results} presents the performance comparison of various actor models under different frameworks. Across five long-horizon information-seeking benchmarks, \ARC{} improves over both ReAct and ReSum for every actor where results are reported, and the gains are most pronounced on the hardest search-heavy settings (GAIA, xBench-DS, BrowseComp, BrowseComp-ZH). 

Two patterns stand out. 
First, the advantage of \ARC{} amplifies on harder, long-horizon tasks, revealing a key insight: the harder the task, the more context must be managed, not merely compressed. As interaction histories become longer and more error-prone, passive summarization struggles to maintain alignment, while active context management enables agents to continually correct and refocus their internal reasoning state.
Second, \ARC{} benefits both small and large actors: it substantially lifts weaker actors (e.g., Qwen2.5-7B-Instruct, Qwen3-14B) while still improving strong ones (e.g., Qwen3-32B, DeepSeek-v3.2), implying that context management is complementary to model scaling rather than a substitute. 

Overall, the results support the view that long-horizon failures are often driven by context degradation, and that reflection-driven reorganization helps prevent error accumulation and unproductive loops in multi-step search.

\begin{table*}[t]
\centering
\small
\setlength{\tabcolsep}{2.0pt}
\renewcommand{\arraystretch}{1.1}
\begin{tabular}{llcccccccccc}
\toprule
\textbf{Models} & \textbf{Framework}
& \multicolumn{2}{c}{\textbf{HotpotQA}}
& \multicolumn{2}{c}{\textbf{GAIA}}
& \multicolumn{2}{c}{\textbf{xBench-DS}}
& \multicolumn{2}{c}{\textbf{BrowseComp}}
& \multicolumn{2}{c}{\textbf{BrowseComp-ZH}} \\
\cmidrule(lr){3-4}\cmidrule(lr){5-6}\cmidrule(lr){7-8}\cmidrule(lr){9-10}
\cmidrule(lr){11-12}
& & Pass@1 & Pass@3
  & Pass@1 & Pass@3
  & Pass@1 & Pass@3
  & Pass@1 & Pass@3
  & Pass@1 & Pass@3 \\
\midrule

\multirow{3}{*}{\makecell[l]{Qwen2.5-7B-Instruct}}
& ReAct & 52.3 & 72.3 & \underline{16.0} & 28.9 & 20.6 & 36.0 & 0.7 & 1.7 & 4.7 & \underline{11.0} \\
& ReSum & \underline{54.8} & \underline{74.4} & 14.7 & \underline{30.2} & \underline{21.0} & \underline{42.0} & \underline{1.0} & \underline{2.0} & \underline{4.8} & 10.7 \\
& \textbf{ARC} & \textbf{66.7} & \textbf{75.9} & \textbf{24.6} & \textbf{38.0} & \textbf{34.3} & \textbf{57.0} & \textbf{2.0} & \textbf{3.2} & \textbf{13.1} & \textbf{25.3} \\
\midrule

\multirow{3}{*}{\makecell[l]{Qwen2.5-32B-Instruct}}
& ReAct & 65.5 & 78.5 & \underline{28.5} & 40.9 & 33.3 & 51.0 & 1.0 & 2.7 & 7.1 & 14.5 \\
& ReSum & \underline{68.6} & \textbf{81.0} & 27.3 & \underline{42.2} & \underline{36.3} & \underline{55.0} & \underline{1.2} & \underline{3.0} & \underline{7.4} & \underline{16.6} \\
& \textbf{ARC} & \textbf{71.3} & \underline{80.5} & \textbf{34.9} & \textbf{50.6} & \textbf{40.7} & \textbf{64.0} & \textbf{3.0} & \textbf{5.8} & \textbf{18.0} & \textbf{33.2} \\
\midrule

\multirow{3}{*}{\makecell[l]{Qwen3-14B}}
& ReAct & 58.3 & 73.4 & 26.5 & 44.6 & \underline{32.6} & 50.0 & \underline{0.8} & \underline{2.0} & 8.5 & 15.9 \\
& ReSum & \underline{59.0} & \underline{74.4} & \underline{31.3} & \underline{49.4} & 32.0 & \underline{56.0} & 0.7 & 1.4 & \underline{10.0} & \underline{18.7} \\
& \textbf{ARC} & \textbf{74.6} & \textbf{82.8} & \textbf{41.0} & \textbf{56.6} & \textbf{46.0} & \textbf{66.0} & \textbf{4.0} & \textbf{6.8} & \textbf{17.9} & \textbf{32.2} \\
\midrule

\multirow{3}{*}{\makecell[l]{Qwen3-32B}}
& ReAct & 58.6 & 73.4 & \underline{29.7} & 48.2 & \underline{33.7} & \underline{52.0} & 1.6 & 3.0 & \underline{11.1} & 20.7 \\
& ReSum & \underline{60.7} & \underline{76.0} & 27.3 & \underline{49.0} & 30.0 & 48.0 & \underline{2.0} & \underline{4.1} & 10.7 & \underline{21.5} \\
& \textbf{ARC} & \textbf{74.0} & \textbf{81.8} & \textbf{44.6} & \textbf{57.8} & \textbf{44.7} & \textbf{68.0} & \textbf{4.4} & \textbf{8.2} & \textbf{19.7} & \textbf{32.8} \\
\midrule

\multirow{3}{*}{\makecell[l]{DeepSeek-v3.2}}
& ReAct & 79.6 & \textbf{88.5} & 63.7 & 78.3 & 63.7 & 77.0 & 16.8 & 31.5 & 37.7 & 56.4 \\
& ReSum & \textbf{80.7} & 85.7 & \underline{67.1} & \underline{79.5} & \underline{66.3} & \textbf{82.0} & \underline{22.2} & \underline{35.6} & \underline{45.7} & \underline{60.5} \\
& \textbf{ARC} & \underline{79.8} & \underline{88.1} & \textbf{69.1} & \textbf{81.0} & \textbf{69.7} & \textbf{82.0} & \textbf{26.6} & \textbf{39.7} & \textbf{51.7} & \textbf{63.8} \\

\bottomrule
\end{tabular}
\caption{Main results on long-horizon information-seeking benchmarks (Pass@1 / Pass@3).
\textbf{ReAct} appends raw interaction history, \textbf{ReSum} applies static/periodic summarization, and \ARC{} performs always-on incremental summarization with triggered reflection-driven reorganization of interaction memory and checklist.}
\label{tab:main_results}
\end{table*}

\subsection{Ablation Studies}
\label{sec:ablation}

We perform ablation studies within the \ARC{} framework to isolate the contributions of the active incremental memory construction, checklist-based control, and reflection-driven memory revision.

\textbf{Summary} maintains the interaction memory using incremental online summarization only.
No reflection is applied, and neither the memory nor the checklist is revised after initialization.

\textbf{Summary + Checklist} allows reflection to update the checklist as a control signal, while the interaction memory remains purely incremental and is never revised.

\textbf{Reflection + Checklist-Only} triggers reflection when execution stalls and permits checklist revision, but explicitly disables any modification to the interaction memory, isolating the effect of control adjustment without memory repair.

\textbf{ARC (Full)} combines incremental interaction memory construction with triggered reflection that jointly revises both the interaction memory and the checklist, enabling explicit repair and reorganization of past context.

All variants share the same actor model (Qwen2.5-32B-Instruct).
Performance differences therefore isolate the impact of reflection-driven interaction memory revision beyond incremental summarization and checklist-based control alone.

\begin{table}[h]
\centering
\small
\setlength{\tabcolsep}{3pt}
\begin{tabular}{lccc}
\toprule
Method & GAIA & xBench-DS & BrowseComp-ZH\\
\midrule
Summary & 32.1 & \underline{38.4} & 14.4\\
Summary+Checklist & \underline{33.0} & 37.6 & \underline{15.0} \\
Reflection+Checklist & 26.2 & 34.3 & 6.9 \\
\midrule
\textbf{ARC (Full)} & \textbf{34.9} & \textbf{40.7} & \textbf{18.0} \\
\bottomrule
\end{tabular}
\caption{Ablation study of different context management strategies within the \ARC{} framework on GAIA, xBench-DeepSearch and BrowseComp-ZH.}

\label{tab:ablation}
\end{table}

The ablation results in Table~\ref{tab:ablation} clarify the roles of different components in \ARC{}.

Incremental summarization provides a clear and intuitive benefit by stabilizing context length and preserving task-relevant evidence.
However, checklist-based reflection alone yields only marginal or inconsistent gains, indicating that control signals without memory revision are insufficient.

The strongest improvements arise when reflection operates over a compact and actively managed interaction memory.
\ARC{} (Full) consistently outperforms all ablations, especially on more challenging benchmarks, suggesting that reflection is most effective when applied to a distilled context rather than raw or passively accumulated history.
These results support the view that the checklist is most valuable as part of a jointly revisable internal state, rather than as an isolated planning artifact, and that effective long-horizon reasoning requires combining compression with explicit memory repair.

\subsection{Effect of Different Context Managers}
\label{sec:cm_effect}

\ARC{} decouples context management from action execution, allowing the Context Manager (CM) to be instantiated by different models.
To examine the impact of CM choice, we fix the Actor and execution loop, and vary only the model responsible for interaction memory construction and reflection.

Results in Table~\ref{tab:cm_models} show that CM choice consistently affects performance, especially on long-horizon benchmarks such as GAIA and xBench-DS.
While stronger general-purpose models yield better results, a trained ARC-CM based on Qwen3-14B outperforms substantially larger untrained models such as GPT-OSS-120B across all benchmarks.
This result indicates that \ARC{} is not merely an emergent property of large models, but a capability that can be explicitly learned through supervision.

Importantly, the trained CM achieves stronger performance while operating at significantly lower model capacity, reducing summarization and reflection costs during execution.
These findings suggest that actively learned context management can raise the effective reasoning ceiling of the Actor, enabling more reliable long-horizon behavior without increasing the Actor’s own model size.

\begin{table}[t]
\centering
\small
\setlength{\tabcolsep}{3.5pt}
\begin{tabular}{lccc}
\toprule
Context Manager & HotpotQA & GAIA & xBench-DS \\
\midrule
Qwen3-32B & 72.1 & 35.1 & 39.0 \\
GPT-OSS-120B & 74.6 & 34.9 & 40.7 \\
DeepSeek-v3.2 & \underline{75.4} & \textbf{48.4} & \textbf{52.0} \\
\midrule
\textbf{ARC-CM} & \textbf{76.1} & \underline{42.6} & \underline{46.3}\\
\bottomrule
\end{tabular}
\caption{Effect of different Context Managers (CMs) in \ARC{}.
The Actor and execution pipeline are fixed, while only the CM is varied.
A trained ARC-CM based on Qwen3-14B outperforms larger untrained models (e.g., GPT-OSS-120B), indicating that \ARC{} is a learnable capability rather than a byproduct of model scale.}
\label{tab:cm_models}
\end{table}

\subsection{Management Frequency: Why Per-Turn Active Context Management}
\label{sec:analysis_summary}

To analyze the role of active context management frequency, we vary when summarization and reflection checks are performed within ARC while keeping the Actor, Context Manager, tools, and context budget fixed.
Specifically, we compare (i) step-by-step management that performs summarization and reflection checks at every turn, (ii) delayed management that updates context every 3 or 5 turns, and (iii) budget-triggered management that activates only when the context length exceeds a predefined token threshold (8k, 16k, or 32k).
All settings are evaluated under the same interaction limits to isolate the effect of context management frequency.

\begin{table}[h]
\centering
\small
\setlength{\tabcolsep}{8pt}
\renewcommand{\arraystretch}{1.0}
\begin{tabular}{lcc}
\toprule
\textbf{Strategy} & \textbf{Setting} & \textbf{Accuracy (\%)} \\
\midrule
Step-by-step & Every turn & \textbf{31.2} \\
\midrule
Delayed & Every 3 turns & 26.5 \\
Delayed & Every 5 turns & 24.5 \\
\midrule
Budget-triggered & 8k tokens & 27.1 \\
Budget-triggered & 16k tokens & 24.4 \\
Budget-triggered & 32k tokens & 24.6 \\
\bottomrule
\end{tabular}
\caption{Accuracy is reported as the average performance over GAIA, xBench-DeepSearch, and BrowseComp.
Per-turn summarization and reflection checks outperform delayed and budget-triggered strategies.}
\label{tab:compression-accuracy}
\end{table}

Table~\ref{tab:compression-accuracy} shows that step-by-step context management consistently outperforms delayed and budget-triggered strategies.
A key advantage of per-turn management is that each update preserves complete semantic information from the most recent interaction, before any abstraction or omission accumulates across multiple steps.

In contrast, delayed summarization must compress several interactions at once, increasing the risk that intermediate reasoning, failed attempts, or subtle evidence is lost or conflated.
Budget-triggered strategies are similarly reactive: by the time compression is applied, misleading assumptions or unproductive patterns may already have shaped subsequent decisions.

These results indicate that context management in long-horizon information seeking is not merely about reducing length, but about maintaining a semantically faithful and task-aligned internal reasoning state.
By incrementally summarizing each interaction with full semantic coverage and performing lightweight reflection checks at every turn, \ARC{} prevents early deviations from compounding and enables timely correction.
This motivates our design choice of always-on, per-turn context management.

%% file: Body/5.conclusion.tex
\section{Conclusion}

In this work, we study context management as a central challenge in long-horizon information-seeking agents.
We argue that failures in deep search often stem not from insufficient reasoning capability, but from how interaction history is compressed, organized, and reused over time.

We propose \textbf{ARC}, an active and reflection-driven context management framework that treats context as a dynamically managed internal state rather than a static summary or an append-only record.
\ARC{} introduces an always-on, incremental interaction memory together with selectively triggered reflection, enabling agents to actively monitor, revise, and realign their internal reasoning state during execution.
Importantly, context management in \ARC{} is decoupled from action generation and implemented by a dedicated, learnable \textbf{Context Manager}, making it reusable across different actors and tasks.

Across multiple long-horizon benchmarks, \ARC{} consistently improves robustness and reliability compared to raw history concatenation and passive summarization baselines.
Further experiments show that effective context management is a learnable capability: a trained Context Manager can outperform much larger untrained models, while reducing summarization cost and raising the effective reasoning ceiling of the actor.

Taken together, our results suggest that context management should be viewed as an active, first-class component in agent design.
By moving beyond passive compression toward active and reflection-driven state management, \ARC{} offers a principled and scalable approach to supporting long-horizon reasoning in information-seeking agents.

Additional analyses on interaction efficiency and token efficiency are provided in Appendix~\ref{app:analysis}. 
Appendix~\ref{app:case} presents qualitative case studies that demonstrate how \ARC{} corrects erroneous assumptions and reorganizes its interaction memory in practice.

%% file: Body/limitations.tex
\section*{Limitations}

Despite its effectiveness, \ARC{} has several limitations.

First, \ARC{} introduces an explicit context management component, which inevitably incurs additional computational and system overhead compared to single-model agent designs.
Although the Context Manager is decoupled from action execution and can be implemented efficiently, active context monitoring and revision introduce extra cost that may be undesirable in latency- or resource-constrained settings.

Second, our evaluation focuses on long-horizon information-seeking and search-based tasks.
While these settings capture a broad and practically important class of agent behaviors, the applicability of reflection-driven context management to other agentic scenarios remains to be explored.

Third, training the Context Manager relies on curated trajectories that exhibit desirable summarization and reflection behaviors.
Although supervised fine-tuning enables \ARC{} to internalize effective context management strategies, the dependence on annotated or filtered data may limit scalability.
Exploring more data-efficient or self-supervised training paradigms is an important direction for future work.

We leave these directions for future research.

%% file: Body/appendix.tex
\appendix

\section{Training Details}
\label{app:training}

\subsection{Training Objective}

We treat the Context Manager (CM) as a learnable module dedicated to \emph{active context management}, encompassing both incremental summarization and reflection-driven context reorganization.
The CM is trained via supervised fine-tuning (SFT) to predict updated internal states—specifically, the interaction memory and checklist—conditioned on the task query, the previous internal context state, and the most recent interaction.

Formally, given a query $q$, a prior context state $C_{t-1}$, and the latest interaction $I_t = (r_t, a_t, o_t)$, the CM learns to produce an updated context state $C_t = (\text{memory}_t, \text{checklist}_t)$.
This formulation allows the CM to internalize both routine summarization and reflection-driven repair behaviors within a single forward pass, without direct access to future actions or environment feedback.
Importantly, the CM is trained independently of action generation, enabling context construction and revision to be learned as a standalone capability rather than an emergent byproduct of policy optimization.

\subsection{Data Construction}

Training data is constructed from agent trajectories collected across a diverse set of long-horizon, information-seeking benchmarks, including
2WikiMultiHopQA~\citep{xanh2020_2wikimultihop},
Bamboogle~\citep{press2023measuringnarrowingcompositionalitygap},
WebWalkerQA~\citep{wu-etal-2025-webwalker},
MuSiQue~\citep{trivedi2021musique},
and the training split of HotpotQA~\citep{yang2018hotpotqa}.

For each benchmark, we deploy agents in an iterative search-and-reasoning setting to collect full execution trajectories.
Each trajectory consists of a sequence of interactions $I_t = (r_t, a_t, o_t)$, together with the corresponding interaction memory and checklist states generated during execution.
These intermediate context states serve as supervision targets for training the CM to update internal context representations step by step.

To ensure data quality, we filter out trajectories with malformed context updates, invalid checklist transitions, or unrecoverable execution failures.
Only trajectories exhibiting coherent incremental summarization and valid reflection behavior are retained.
The resulting dataset provides dense supervision for learning context construction, maintenance, and repair in long-horizon information-seeking scenarios.
All trajectories are generated using an agent with strong context management capabilities, ensuring coherent context updates and reliable reflection behavior for supervision.

\subsection{Training Setup}

We initialize the Context Manager from Qwen3-14B and fine-tune all model parameters using full-parameter supervised fine-tuning.
Training is conducted on two nodes with a total of 16 NVIDIA H100 GPUs.
The specific training parameters are shown in Table~\ref{tab:training_config}

\begin{table}[h]
\centering
\small
\setlength{\tabcolsep}{6pt}
\renewcommand{\arraystretch}{1.1}
\begin{tabular}{ll}
\toprule
\textbf{Setting} & \textbf{Value} \\
\midrule
Max sequence length & 16,000 \\
Per-device batch size & 2 \\
Gradient accumulation & 2 \\
Learning rate & $5\times10^{-6}$ \\
LR scheduler & Cosine with 10\% warmup \\
Precision & bfloat16 \\
Parallelism & Sequence parallelism (size = 2) \\
Memory optimization & DeepSpeed ZeRO-3 \\
\bottomrule
\end{tabular}
\caption{Training configuration for the Context Manager.}
\label{tab:training_config}
\end{table}

\section{Main Experiment Evaluation Settings}
\label{app:eval}

This appendix describes the evaluation settings used in the main experiments.

All agents are evaluated in an information-seeking environment equipped with two tools: \textbf{search} and \textbf{visit}.
The \textbf{search} tool is used to issue search queries and retrieve candidate webpages, while the \textbf{visit} tool allows the agent to inspect the content of selected pages.
Agents may invoke these tools iteratively throughout execution.

To ensure a consistent and fair comparison across methods, all agents operate under the same resource constraints.
Specifically, the maximum context window is limited to \textbf{32k tokens}, and the maximum number of interaction turns is capped at \textbf{80} per episode.
An episode terminates early if the agent produces a confident final answer before reaching the turn limit.
When the maximum number of interaction turns is reached or the context window limit is exceeded, the agent is forced to produce a final answer by consolidating the available information.

All baselines and \ARC{} variants share the same tool interface, context budget, and termination criteria.
Differences in performance therefore reflect variations in context management strategies rather than disparities in tool access or computational resources.

\section{Analysis}
\label{app:analysis}
\subsection{Interaction Efficiency}
\label{sec:analysis_steps}

We analyze interaction efficiency by examining how task accuracy evolves as the maximum number of allowed interaction steps increases.
For each method (\textbf{ReAct}, \textbf{ReSum}, and \textbf{\ARC{}}), we progressively relax the interaction step limit and measure the corresponding task success rate.

\begin{figure}[h]
    \centering
    %\hspace*{-0.07\columnwidth}
    \includegraphics[width=1.0\columnwidth]{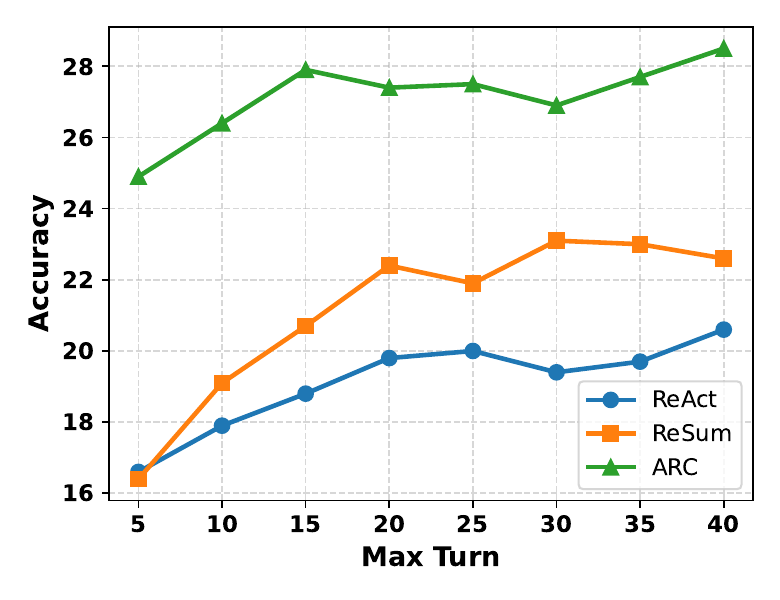}
    \caption{Interaction efficiency under varying maximum interaction turns.
    Accuracy denotes the average performance on xBench-DeepSearch and BrowseComp-ZH, reflecting how effectively each method utilizes additional interaction budget.}
    \label{fig:max_turn}
\end{figure}

Figure~\ref{fig:max_turn} plots task accuracy as a function of the maximum allowed interaction turns.
As the interaction budget increases, all methods benefit from additional exploration, but their improvement patterns differ substantially.

ReAct exhibits steady but limited gains, indicating that simply accumulating raw interaction history does not effectively translate additional steps into better decisions.
ReSum improves more rapidly at early stages, but its performance plateaus as static summaries begin to absorb unproductive exploration and early errors.

In contrast, \ARC{} consistently achieves higher accuracy across all interaction budgets and continues to improve as more turns are allowed.
This suggests that \ARC{} is more effective at converting additional interaction steps into useful progress.
By actively revising its interaction memory through reflection, \ARC{} prevents repeated exploration of unproductive paths and redirects search more efficiently.
As a result, \ARC{} reaches stronger performance with fewer wasted interactions, demonstrating superior interaction efficiency in long-horizon information-seeking settings.
\subsection{Token Efficiency}

We analyze token efficiency by tracking how context token usage grows as interaction length increases.
All results are reported using the average interaction length observed on the BrowseComp benchmark, which reflects realistic long-horizon information-seeking trajectories.

\begin{figure}[h]
    \centering
    %\hspace*{-0.07\columnwidth}
    \includegraphics[width=1.0\columnwidth]{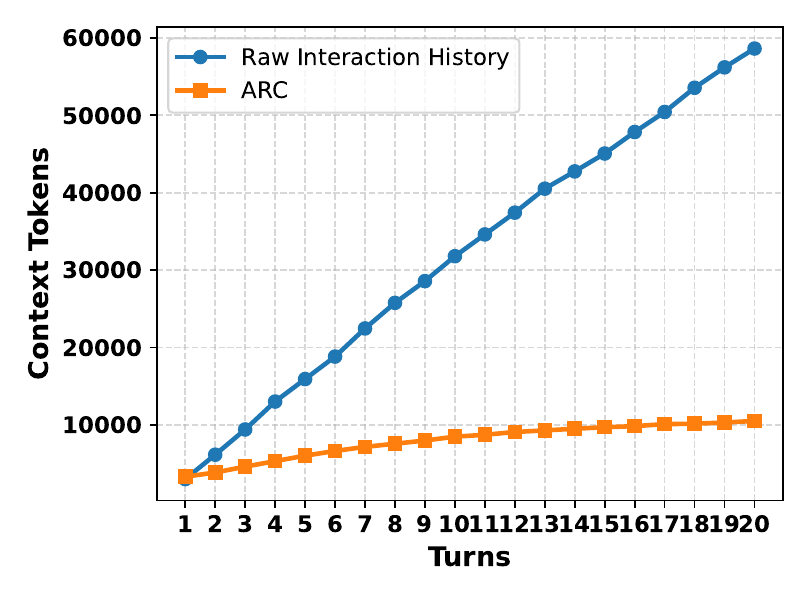}
    \caption{Token efficiency on BrowseComp. Context token usage is averaged over trajectories as interaction length increases. \ARC{} maintains a compact interaction memory, while raw interaction history grows nearly linearly.}
    \label{fig:tokens}
\end{figure}

Figure~\ref{fig:tokens} compares \ARC{} with a baseline that appends the raw interaction history at each turn.
As interaction length increases, raw history concatenation exhibits near-linear growth in context tokens, quickly exhausting the available context budget.
In contrast, \ARC{} maintains a compact interaction memory whose token usage grows slowly and stabilizes over time.

This gap arises because \ARC{} incrementally summarizes interactions and actively reorganizes its internal memory, preventing redundant or uninformative content from accumulating.
By contrast, raw history concatenation preserves all intermediate reasoning, actions, and observations, including failed attempts and repeated patterns, leading to rapid context inflation.

These results demonstrate that \ARC{} not only improves reasoning robustness, but also substantially reduces cumulative context usage.
By actively managing context as a dynamic internal state rather than an append-only history, \ARC{} enables long-horizon interaction under fixed context budgets, making it more practical for real-world information-seeking settings.

Taken together, the interaction and token efficiency analyses reveal a consistent pattern: effective long-horizon reasoning depends not on accumulating more interactions or context, but on how actively the internal state is managed.
\ARC{} achieves higher task performance with fewer redundant interactions and substantially lower context usage by continuously summarizing and revising its interaction memory.
By preventing unproductive exploration and eliminating irrelevant or outdated information early, \ARC{} converts each interaction into more meaningful progress while maintaining a compact and semantically faithful context.
These results further support our view that context management should be treated as an active, first-class component of agent design, rather than a passive byproduct of history accumulation.

\section{Case Study}
\label{app:case}

To qualitatively analyze how \ARC{} manages long-horizon context, we present a representative case study.

In the provided case, the model becomes trapped in a local search loop centered on "Susunu! Denpa Shōnen".
This failure arises because the checklist prematurely elevates the cue “an Asian man being exploited for three months” to a dominant constraint, causing subsequent reasoning steps to focus on validating a single candidate rather than reassessing alternative explanations.

During reflection, the checklist is treated as an editable execution-time state rather than a fixed plan.
Revision is triggered when the agent detects confirmation-driven behavior, such as repeatedly searching for evidence that reinforces the same hypothesis without resolving key inconsistencies.
% \begin{figure*}[t]
% \centering

\begin{tcolorbox}[
    breakable,
    colback=white,
    colframe=black,
    boxrule=0.6pt,
    arc=2pt,
    left=6pt,
    right=6pt,
    top=6pt,
    bottom=6pt
]
\textbf{Query.}  
\emph{I am looking for a TV show aired sometime between 1990 and 2019. I can't remember many details. But I do remember these events happening in this order:  - One of the characters is standing near the bathroom and also near another character with the same first name. - Someone exploits an Asian man for three months. - Someone notes that the vending machines are full of meats.  What is the name of this TV show?}

\textbf{Labeled Answer: Community}\\
\vspace{0.6em}

\textbf{[Turn 1]}

\textbf{[Checklist]} \\
1. Identify unique plot points \\
2. Search for shows with exploitation plot \\
3. Cross-reference with vending machine detail \\
4. Confirm character name coincidence \\

\textbf{Reasoning:}
The clues are quite specific but require finding a match across multiple plot points. First, I need to start by searching for TV shows that involve the exploitation of an Asian man for three months, as this seems the most unique and searchable detail.\newline{}\textbf{Action:}  Search \\

\textbf{[Turn 2]}

\textbf{[Checklist]} \\
...

\textbf{[Last Interaction]} \\
...

\textbf{Reasoning:}
The previous search focused on exploitation plots, but the results were not specific enough. The mention of an Asian man exploited for three months in a TV show might refer to the infamous Japanese reality show 'Susunu! Denpa Shōnen' (also known as 'The Denpa Shōnen' or 'The Truman Show-like' experiment), where Nasubi was exploited. However, the task requires a show aired between 1990 and 2019, and the plot points include a character standing near a bathroom with another character of the same name and vending machines full of meats. I need to verify if 'Susunu! Denpa Shōnen' matches all details, especially the vending machine and character name coincidence, which are key to narrowing this down.

\textbf{Action:} Search\\

\textbf{[Turn 3]}

\textbf{[Checklist]} \\
...

\textbf{[Interaction Memory]}

[Turn 1] In the first turn, the assistant aimed to identify a TV show by focusing on the unique detail of an Asian man being exploited for three months. A search was conducted to find matching plot points, prioritizing specificity for better results.\\

\textbf{[Last Interaction]} \\
...\\

\textbf{[Turn 4]}

\textbf{[Checklist]} \\
...

\textbf{[Interaction Memory]}

[Turn 1][Turn 2]

\textbf{[Last Interaction]} \\
...\\

\textbf{[Turn 5]}

\textbf{[Checklist]} \\
...

\textbf{[Interaction Memory]}

[Turn 1][Turn 2][Turn 3]

\textbf{[Last Interaction]} \\
...\\

\textbf{[Turn 6]}

\textbf{[Checklist]} \\
...

\textbf{[Interaction Memory]}

[Turn 1][Turn 2][Turn 3][Turn 4]

\textbf{[Last Interaction]} \\
...\\

\textbf{[Turn 7]}

\textbf{[Checklist]} \\
...

\textbf{[Interaction Memory]}

[Turn 1][Turn 2][Turn 3][Turn 4]

[Turn 5] Turn 5 aimed to confirm if Susunu! Denpa Shōnen includes the vending machine meats detail by combining the show's name with that specific element in a search. The action targeted this unique feature to verify alignment with all described plot points. The strategy remained focused on precise narrative cross-checking for conclusive identification.

\textbf{[Last Interaction]} \\
...\\

\textbf{[Turn 8]}

\textbf{[Reflection]}\\
The previous reasoning appears to be overly anchored to a single hypothesis centered on reality TV, driven primarily by the cue describing the exploitation of an Asian man. While salient, this cue may be noisy or conflated with external memories, and repeated searches have focused on validating the same candidate without resolving inconsistencies in other plot elements.

To progress, it is necessary to re-evaluate the relative reliability of the remaining cues, particularly the vending machine filled with meat and the bathroom scene involving same-name characters, which are more characteristic of scripted comedic settings. I should broaden the search space beyond reality shows and explore ensemble sitcoms that better align with these narrative and stylistic features.

\textbf{[Checklist](new)} \\
1. Reopen the candidate space beyond reality TV to include scripted ensemble sitcoms consistent with absurd visual gags.

2. Anchor retrieval on the most identifiable visual cue (the vending machine filled with meat) and require explicit evidence.

3. Verify the bathroom and same-name cues, and assess whether the exploitation cue could reflect a misremembered subplot or joke.

4. If repeated searches fail to verify key cues, discard the current anchor candidate and restart with revised cue weights.

\textbf{[Interaction Memory])(new)}

[Turn 1-5] The agent has repeatedly focused on Susunu! Denpa Shōnen as the main candidate and conducted multiple searches to cross-check different clues, including the exploitation detail, the vending machine scene, and the bathroom-related interaction. However, none of these searches produced clear or consistent evidence that directly supports this candidate.
After several rounds of verification without concrete matches, it becomes unlikely that Susunu! Denpa Shōnen is the correct answer.

[Turn 6]

\textbf{[Last Interaction]} \\
...\\

...... \\

\textbf{[Turn 31]}

\textbf{[Checklist]} \\
...

\textbf{[Interaction Memory]}\\
...

\textbf{[Last Interaction]} \\
...

\textbf{Reasoning:} After aligning the clues with specific scenes and characters, the evidence converges within a single show. The detail that the vending machines are filled with meat directly corresponds to a scene in Community, where Abed Nadir explicitly notices and points out the abnormal contents of the vending machine while speaking with Jeff Winger, matching the show’s characteristic use of concrete, absurd visual gags. The description of a same-name interaction occurring near a bathroom is also consistent with Community’s frequent use of campus public spaces and dialogue-driven name-based humor. Finally, the clue describing the exploitation of an Asian man over a period of time aligns at the character level with Ben Chang, whose role in the series repeatedly involves being used and manipulated by the school system and other characters. These character- and scene-level correspondences can all be satisfied within the same narrative context, leading to a consistent identification of the show as Community.

\textbf{Answer:} Community.

\end{tcolorbox}